\documentclass[10pt, a4paper]{article}

\usepackage{lrec-coling2024} 

\usepackage{graphicx} 
\usepackage{enumerate}
\usepackage{enumitem}
\usepackage{amsmath}
\usepackage{booktabs}
\usepackage{multirow}
\usepackage{floatrow}
\usepackage[label font=bf,labelformat=simple]{subfig}
\usepackage{array}
\usepackage{booktabs}

\newenvironment{shrinkeq}[1]
{ \bgroup
\addtolength\abovedisplayskip{#1}
\addtolength\belowdisplayskip{#1}}
{\egroup\ignorespacesafterend}

\title{Towards Generalizable and Faithful Logic Reasoning over Natural Language via Resolution Refutation}

\name{Zhouhao Sun$^1$, Xiao Ding$^1$\sthanks{*Corresponding Author}, Li Du$^2$, Bibo Cai$^1$, Jinglong Gao$^1$, Ting Liu$^1$, Qin Bing$^1$ } 

\address{$^1$Research Center for Social Computing and Information Retrieval\\
         Harbin Institute of Technology, China\\
         $^2$Beijing Academy of Artificial Intelligence, Beijing, China \\
         \{zhsun, xding, bbcai, jlgao, tliu, bqin\}@ir.hit.edu.cn\\
         duli@baai.ac.cn\\}

\abstract{
Large language models (LLMs) have achieved significant performance in various natural language reasoning tasks. However, they still struggle with performing first-order logic reasoning over formal logical theories expressed in natural language. This is because the previous LLMs-based reasoning systems have the theoretical incompleteness issue. As a result, it can only address a limited set of simple reasoning problems, which significantly decreases their generalization ability. To address this issue, we propose a novel framework, named \underline{G}eneralizable and \underline{Fai}thful \underline{R}easoner (\underline{GFaiR}), which introduces the paradigm of resolution refutation. Resolution refutation has the capability to solve all first-order logic reasoning problems by extending reasoning rules and employing the principle of proof by contradiction, so our system’s completeness can be improved by introducing resolution refutation. Experimental results demonstrate that our system outperforms previous works by achieving state-of-the-art performances in complex scenarios while maintaining performances in simple scenarios. Besides, we observe that GFaiR is faithful to its reasoning process. 
 \\ \newline }

\begin{document}

\maketitleabstract

\section{Introduction}
\begin{figure}[!htb]
    \centering
    \sidesubfloat[]{
    \includegraphics[width=0.95\columnwidth]{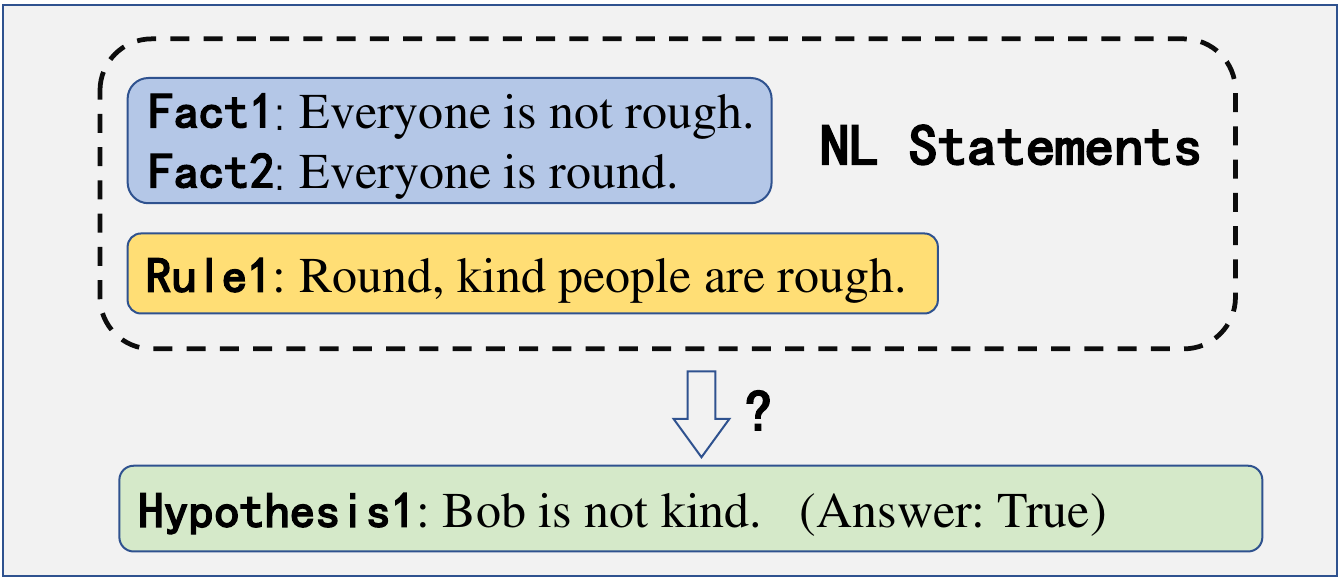}
    \label{fig:subfig_a}
    }
    \quad
    \centering
    \sidesubfloat[]{
    \includegraphics[width=0.95\columnwidth]{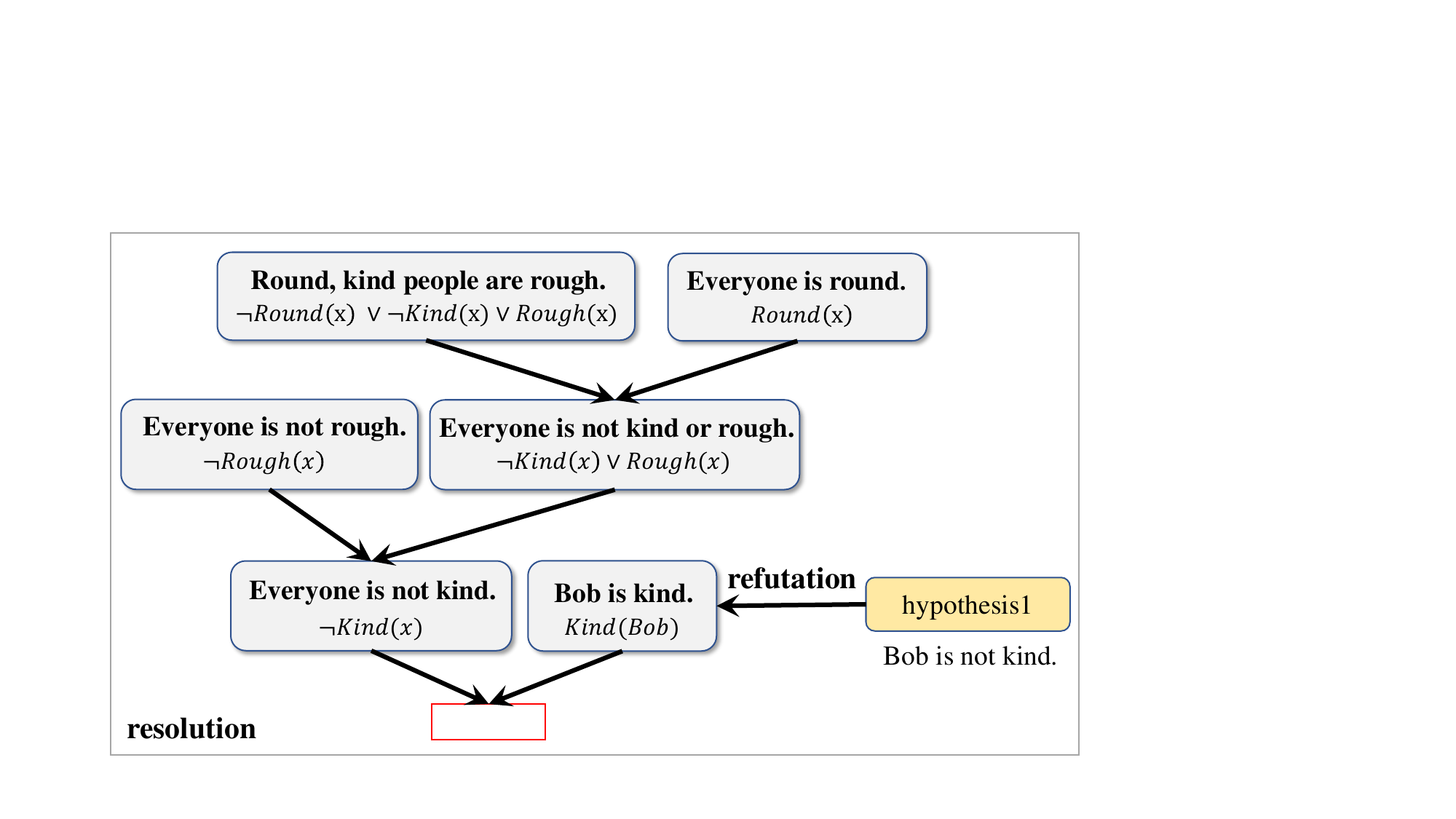}
    \label{fig:subfig_b}
    }
    \caption{(a) Example of an NL Theory and a hypothesis with gold answers. Note that the meaning of these statements are not related to the common sense. (b) For hypothesis 1, the reasoning process using the method of resolution refutation is shown. The process of refutation is reflected from 'hypothesis' to "Bob is kind" and the grey box represents the process of resolution at the natural language level.}
\end{figure}

The natural language-based logical reasoning task requires the model to understand the abstract logical relationships within statements expressed in natural language to deduce a conclusion. For example, as shown in Figure \ref{fig:subfig_a}, the task is to determine the value of the hypothesis (True, False, Unknown) based on a natural language theory (\textbf{NL Theory}) which consists of a set of rules and facts explicitly stated in natural language. This task is increasingly gaining attention \cite{sun2021probabilistic,kazemi-etal-2023-lambada}, as it bridges the natural language with abstract logical thinking, which plays a pivotal role in complex problem-solving and cognitive reasoning.

Recently, transformer-based LLMs have achieved significant performance in various natural language reasoning tasks \cite{wei2022chain,qiao2022reasoning}. Theoretical analyses have also demonstrated that transformers have the potential to perform logical reasoning over formal theories \cite{schlegel-etal-2022-transformers,ijcai2023p375}. However, it still remains challenging for the present LLMs \cite{Liangming-Pan-2023,kazemi-etal-2023-lambada}, even for the State-of-the-Art models including ChatGPT \cite{bang2023multitask}. This is because of the \emph{hallucination} problem \cite{golovneva2023roscoe,ribeiro2023street}, i.e., LLMs may hallucinate incorrect intermediate reasoning steps to draw the final conclusions. As a result, the inference results are not \emph{faithful} to be trusted \cite{lyu2023faithful,creswell2022faithful}. Moreover, if regarding large language models as inference systems, hallucination will affect their \emph{completeness}. A complete inference system means that \textbf{all} the hypotheses with determined labels can be inferred by applying valid reasoning rules contained in the inference system. However, the hallucination problem prevents the LLMs from correctly wielding reasoning rules to draw conclusions, thus leading to incompleteness inference systems.

To reduce hallucination and improve faithfulness for LLMs, previous works mainly enhance the reasoning process of LLMs by a stepwise inference paradigm. According to the direction of reasoning, these works can be divided into two groups. The forward chaining approach \cite{sanyal2022fairr} starts from known rules to check if there exists any rule whose conditions are all satisfied by the given facts, if so, we apply the reasoning rule of forward chaining to derive a new conclusion, this procedure continues until no new conclusions can be drawn or the hypothesis is proved. The backward chaining approach \cite{qu2022interpretable} starts from the hypothesis and reasons in an opposite direction to derive a set of facts that need to be satisfied, then querying if these inferred facts overlap the known facts. By introducing intermediate steps, faithfulness can be improved.

However, the performance of these methods in complex logical reasoning scenarios is still unsatisfying. In some cases, their performance may be lower than using LLMs alone, or even lower than random guesses. This is caused by the inherent deficiency of these methods that the forward or backward reasoning method is \textbf{incomplete}. It means that there will be some hypotheses with determined values that are considered Unknown by the model. As a result, it can only accommodate relatively simple scenarios. Take forward chaining as an example, forward chaining is incomplete because it is capable of reasoning if and only if 'all the conditions of a certain rule can be proven to be true based on known facts' (condition 1). However, In the process of reasoning, there are some exceptional cases where forward chaining cannot reason. For the Hypothesis 1 in Figure \ref{fig:subfig_a}, forward chaining is unable to complete this type of reasoning since the condition of the rule ``kind people'' cannot be proven to be true by the facts. Hence, no conclusions can be drawn and hypothesis 1 will be considered Unknown. For the backward chaining, inference also cannot be made since hypothesis 1 ``not kind'' does not appear at the right hand of the rule. Hence the hypothesis will also be considered Unknown.

Inspired by the logical reasoning methods in the field of symbolic logic, we attempt to introduce a complete logical reasoning paradigm (under first-order logic) resolution refutation \cite{Russellartificial:2010} whose reasoning procedure is not constrained by the condition 1 to improve completeness, and propose a novel reasoning framework GFaiR. Figure \ref{fig:subfig_b} illustrates the reasoning process of our model. For hypothesis 1, by utilizing the reasoning rule of resolution, we can derive 'Everyone is not kind' step by step from the known information by performing resolution at the natural language level. Then by refutation, 'kind' appears in the known information so we can finally prove that hypothesis 1 is True. As a result, the combining of resolution refutation enables the model to handle more complex reasoning scenarios and enhances its generalization ability. Because the process of resolution refutation is complex, so we detail them in Section \ref{sec:background}.

To combine resolution, we need to first select two theories and then utilize a reasoning model to perform resolution over them at the natural language level. However, the previous \cite{sanyal2022fairr} transformers-based selection module only considers selecting which theories are more likely to infer the target hypothesis, without taking into account whether these two theories are logically related. This leads to scenarios where the selected theories are completely unrelated, which further causes the failure of resolution and the generation of invalid conclusions which may result in hallucinations. As a result, we use a validity contrastive loss-based verifier to distinguish valid conditions from illogical statements. This ensures that a valid conclusion can be drawn from the selected theories through logical reasoning, thereby providing guarantees for resolution and improving faithfulness by reducing hallucinations. 

We validate our method on the widely adopted Ruletaker dataset and a more challenging Hard Ruletaker dataset, as well as the natural language satisfiability (NLSAT). Experimental results show that our approach is faithful to its reasoning process and has maintained in-domain inference accuracy, meanwhile demonstrating stronger zero-shot generalization performances\footnote{The source code of GFaiR has been made available at https://github.com/spirit-moon-fly/GFaiR.}. 

\section{Background}
\label{sec:background}

\textbf{Natural Language Reasoning with First-Order Logic} We follow the task definition proposed by \citet{han2022folio}. Given a hypothesis $H$ and an NL Theory $NLT$ (including a series of facts and rules expressed in natural language) without contradiction, the goal is to determine the value of $H$: True, False, or Unknown. Note that $NLT$ and $H$ are annotated with parallel FOL (first-order logic) Theory and FOL hypothesis, and the value is determined by the FOL reasoning result of the FOL Theory and FOL hypothesis. If the value is True or False, it is expected to give a reasoning process, which consists of a series of reasoning steps $\left(p_{1}, p_{2}, ..., p_{n}\right)$, and each reasoning step $p_i$ includes selected rules or facts $s_i$ along with reasoning conclusion $c_i$.

\noindent
\textbf{Resolution Refutation} Resolution refutation \cite{nawaz2019survey} is a commonly used and complete reasoning method under first-order logic, i.e. for a hypothesis whose label is True or False under the semantics of full FOL, applying the reasoning method of resolution refutation can infer the label of the hypothesis. Let $F$ be the FOL formula set of the given premises, and $Q$ be the hypothesis, then the process of proving that $Q$ is True by resolution refutation is shown as follows:
\begin{itemize}[itemsep=0pt,topsep=0pt,parsep=0pt,leftmargin=*]
  \item [1)] 
  Negate $Q$ to get $\neg Q$, and merge it into the formula set $F$ to get $\left\{F,\neg Q\right\}$. 
  \item [2)]
  Transform $\left\{F,\neg Q\right\}$ into a clause set in Skolem normal form. (Skolem standardization)
  \item [3)]
  Apply resolution principle \cite{robinson1965machine} to resolve clauses in the clause set, each resolution step generates a resolved clause, which is then added to the clause set. This process is repeated iteratively. If an empty clause is obtained during the resolution step, it indicates a contradiction in the clause set and proves that $Q$ is True.
\end{itemize}

\begin{figure*}[!ht]
    \centering
    \includegraphics[width=1.0\columnwidth]{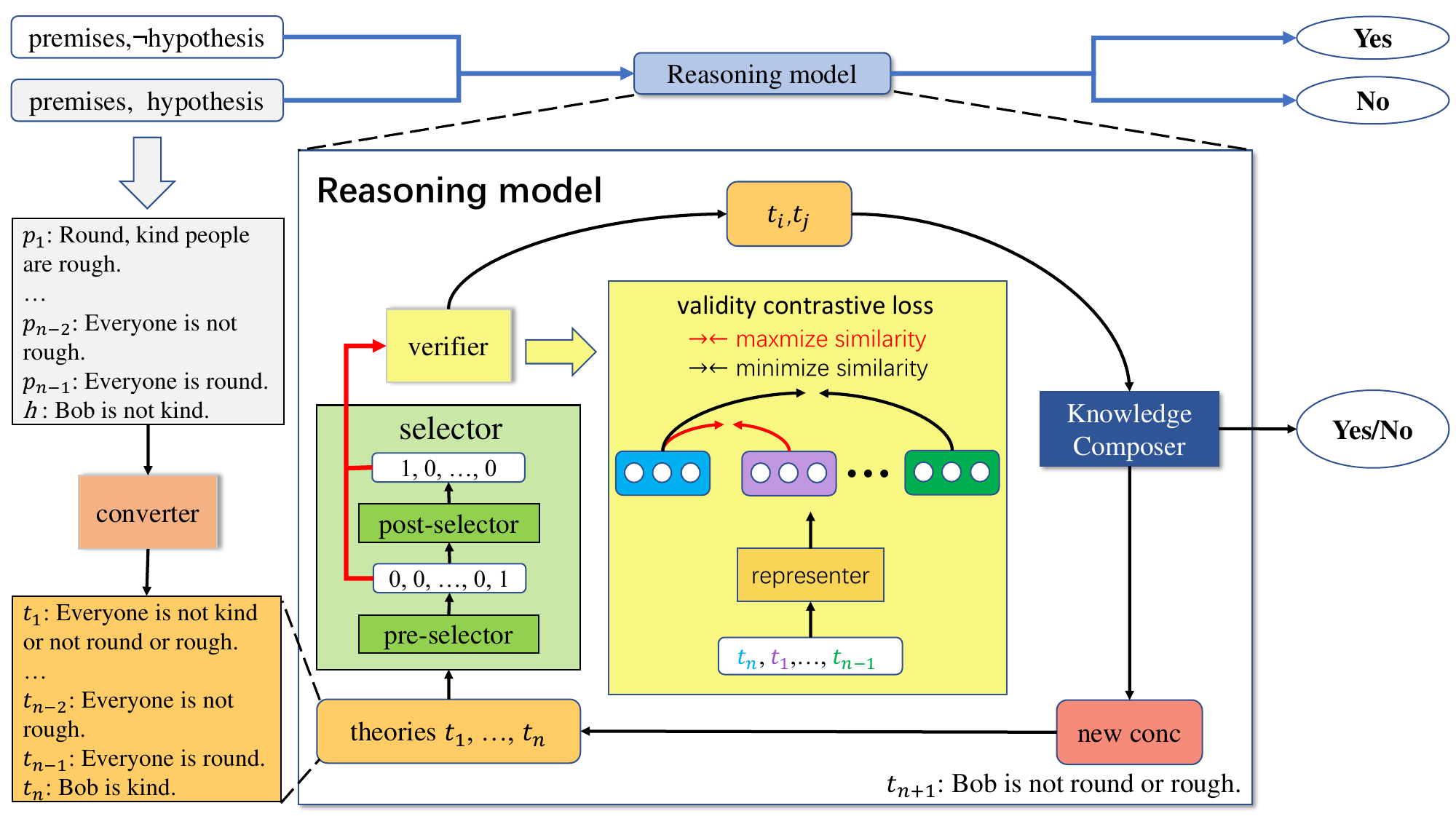}
    \caption{Architecture of GFaiR. We mark the converter in orange, the selector (consisting of pre-selector and post-selector) in green, the verifier with its validity contrastive loss in yellow, and the knowledge composer in blue, respectively.}
    \label{fig:model}
\end{figure*}

The process of proving that $Q$ is False is similar. Therefore, when dealing with our target task, we can determine the value of $H$ by applying a reasoning model to the theory set $T_1$ composed of $NLT$ and $H$ and the theory set $T_2$ composed of $NLT$ and $\neg H$ at the same time, where the reasoning model implicitly performs resolution at the natural language level. If there is no contradiction in the theory set $T_1$ and there is a contradiction in the theory set $T_2$, it proves that $H$ is True. On the contrary, it proves that $H$ is False. if there are no contradictions in two theory sets, $H$ is Unknown. an example of resolution refutation reasoning procedure can be seen in Section \ref{sec:inference}.

\section{Method}
\subsection{Overview}
Although improved faithfulness compared with vanilla LLMs, existing stepwise inference methods based on forward or backward chaining are incomplete, which makes them unable to generalize to complex reasoning scenarios. 

In this paper, we propose a novel reasoning framework GFaiR. As shown in Figure \ref{fig:model}, GFaiR introduces resolution refutation to improve the completeness. 

Specifically, GFaiR is composed of five modules: (1) A converter for augmenting the given NL Theory with the negated hypothesis and to convert the representations of natural language for resolution at the natural language level in the following reasoning process. (2) A pre-selector to select a theory for drawing intermediate conclusions. (3) A post-selector to select another theory by explicitly modeling the relationship between the theory selected by the pre-selector and the remaining ones. (4) A knowledge composer to generate a novel conclusion by applying the resolution rule at the natural language level. (5) A verifier to ensure that a valid conclusion can be drawn from the selected theories through logical reasoning, thereby providing guarantees for resolution and improving faithfulness.

In the following sections, we will first introduce the architecture of GFaiR, and then explain the inference and training procedure of GFaiR.
\vspace{-0.3ex}
\subsection{Architecture}
\vspace{-0.2ex}
Overall, GFaiR is an iterative model where the one-hop intermediate conclusions are generated step-by-step. Our model is shown in Figure \ref{fig:model}. Specifically, we have the following five modules:

\noindent
\textbf{Converter} Given the NL Theory and hypothesis, before directly performing reasoning, we first employ a T5-based converter to automatically convert the hypothesis into its negation form for refutation in the inference process (not reflected in Figure \ref{fig:model}). Additionally, because our knowledge composer mimics the resolution step which cannot deal with existential quantifiers and some implicit logical relationships such as $\rightarrow$, we need a step to convert the implicit logical relationships and existential quantifiers while retaining as much of the original text as possible. The convertor can also perform this step by imitating the Skolem standardization step in resolution refutation, which transforms NL Theory and hypothesis (or its negation form) into natural language representations similar to the Skolem normal form.  
As shown in Figure \ref{fig:model}, the converter converts 'Round, kind people are rough' to 'Everyone is not kind or not round or rough'. The converted NL Theory and hypothesis will be taken as inputs for the following reasoning process. 

\noindent
\textbf{Pre-Selector (Pre-S)} The pre-selector is an XLNET-based \cite{yang2019xlnet} classification model that takes the concatenated theories in the theory set as input (including intermediate conclusions, converted NL Theory and hypothesis), and selects a theory for generating new conclusions in the current iterative steps. Taking theory set $T = \left\{t_1, t_2, ..., t_n\right\}$ in Figure \ref{fig:model} as an example, we concatenate and separate them with the [SEP] token to form the input $\boldsymbol{[}CLS\boldsymbol{]}\ {\boldsymbol{[}t_{i}\ \boldsymbol{[}SEP\boldsymbol{]}\boldsymbol{]}}_n$ (${\boldsymbol{[}\ \boldsymbol{]}}_n$ denotes continued concatenation). The output is a one-dimensional vector denoted as $u$, which is obtained by classifying each [SEP] token embedding via a linear binary classification layer. During iteration, we select the theory in front of the [SEP] token corresponding to the maximum value in the vector $u$. The example in Figure \ref{fig:model} illustrates the selection of $t_n$ based on the highest value in $u$.

\noindent
\textbf{Post-Selector (Post-S)} The post-selector is also an XLNET-based classification model aiming to select another theory based on the theory selected by the pre-selector and the remaining theories. We designed this module to explicitly model the relationship between the theory selected by the pre-selector and the remaining ones. As shown in Figure \ref{fig:model}, $t_n$ is the theory selected in the previous step, and then place $t_n$ at the beginning of the input, while keeping the order of the other theories unchanged and concatenating them after $t_n$. We also use [SEP] token to separate these theories to form the input $\boldsymbol{[}CLS\boldsymbol{]}\ t_{n}\ \boldsymbol{[}SEP\boldsymbol{]}\ {\boldsymbol{[}t_{i}\ \boldsymbol{[}SEP\boldsymbol{]}\boldsymbol{]}}_{n-1}$ . The output is a one-dimensional vector $v$, which is obtained by classifying each [SEP] token embedding (except the first [SEP] token) via a linear binary classification layer. Similar to the pre-selector, the example in Figure \ref{fig:model} illustrates the selection of $t_1$ according to the value in vector $v$.

\noindent
\textbf{Knowledge Composer (KC)} The knowledge composer is a generative transformer T5 that can learn the resolution rule implicitly from data, and apply the learned resolution rule at the natural language level to generate a novel conclusion. As shown in Figure \ref{fig:model}, the input is two theories selected by the pre-selector and post-selector ($t_n$ and $t_1$), and the output $t_{n+1}$ is an intermediate conclusion expressed in natural language, which will be merged into the theory set.

\noindent
\textbf{Verifier} 
The previous \cite{sanyal2022fairr} transformers-based selection module is not accurate enough for resolution refutation, which leads to scenarios where the selected theories are unrelated. This causes the failure of resolution and further the generation of invalid conclusions which may result in hallucinations. As a result, we use a validity contrastive loss-based verifier to verify two theories selected by the pre-selector and post-selector to ensure that a valid conclusion can be drawn from them through logical reasoning, thus providing guarantees for resolution and improving faithfulness by reducing hallucinations. The validity contrastive loss is shown in Figure \ref{fig:model}:

To facilitate explanation, we establish the following definitions: A \textbf{theory pair} $\left(t_i, t_j\right)$ composed of two theories $t_i$ and $t_j$ is \textbf{valid} if and only if a valid conclusion can be drawn from them through logical reasoning. Because our knowledge composer emulates the resolution step at the natural language level to draw intermediate conclusions, the criterion for determining if the theory pair is valid lies in whether the FOL expressions corresponding to these two theories can be used for resolution.

We consider all the theory pairs composed of the theory selected by pre-selector and the remaining ones, and then devise validity contrastive loss by maximizing the cosine similarity of valid theory pairs (pink and blue in Figure \ref{fig:model}) while minimizing the cosine similarity of invalid theory pairs (green and blue in the Figure \ref{fig:model}). Please refer to \ref{sec:train} for the loss function.

When verifying the theory selected by the post-selector $t_k$, the verifier first calculates the cosine similarity between $t_k$ and the theory selected by the pre-selector $t_m$. If the similarity score is above 0 (similarity score is within the range of -1 to 1), the theory pair is deemed valid and selected as input for the knowledge composer. Conversely, it is invalid and the post-selector will select a new theory for verification. This process continues until a theory is selected that can form a valid theory pair with $t_m$.

\subsection{Inference} 
\label{sec:inference}
During inference, the converter first converts the NL Theory and hypothesis into two theory sets represented in natural language similar to the skolem normal form. One of them consists of the NL Theory and hypothesis, the other consists of the NL Theory and the negation of the hypothesis. Then we apply our reasoning model (shown in Figure \ref{fig:model}) to two theory sets separately to infer if there exists a contradiction, which determines the value of the hypothesis (referring to the background for more details). Since our model is a neural network model rather than a symbolic reasoning system, there are accidental conditions that contradiction exists in both theory sets. In such cases, we employ a heuristic approach to determine the value of the hypothesis (according to the number of reasoning steps). Below we will explain how to infer a contradiction in a theory set.

For a specific theory set $T$, the pre-selector first selects a theory $t_i$. Then, under the guidance of the verifier, the post-selector selects a theory $t_j$ that can form a valid theory pair with $t_i$. If it does not exist, stop and conclude that there are no contradictions in the theory set. Conversely, the knowledge composer composes two selected theories to generate a new conclusion. If the conclusion is an empty string (corresponding to the empty clause in the process of resolution refutation), it indicates that there is a contradiction in the theory set and stops the iteration. Otherwise, the newly generated conclusion is placed in the theory set $T$ to participate in the following reasoning process. For the example in Figure \ref{fig:model}, the reasoning model will first derive '$t_{n+1}$: Bob is not round or rough.' by resolving '$t_{1}$: Everyone is not kind or not round or rough.' and '$t_{n}$: Bob is kind.'. Then, by combining $t_{n+1}$ and '$t_{n-2}$ Everyone is not rough.', the model can derive '$t_{n+2}$: Bob is not round.'. Finially, we can derive an empty string from $t_{n+2}$ and '$t_{n-1}$: Everyone is round.', which indicates a contradiction in the theory set and illustrated that the hypothesis is True. Due to the infinite search space of first-order logical reasoning, we design a maximum number of reasoning steps N. When it is reached, we assume that there are no contradictions in the theory set and stop iteration. There may be cases where no theories are selected to form a valid theory pair. For example, the theory set is: \{Bob is kind. Bob is tall. Bob is happy.\}. In this situation, we are unable to derive a valid conclusion based on any theory pair. Therefore, we cannot derive any valid conclusions, and we will halt the search and consider that there are no contradictions in this theory set.

\subsection{Training}
\label{sec:train}
Each component of our model is trained separately. The training data of the converter is every fact and rule in the NL Theory and hypothesis as well as their corresponding natural language representations similar to the skolem normal form (or the negation). The following mainly introduces the training methods of the other four modules.

From Background, we know that the resolution refutation process for proving a hypothesis is True or False involves proving a theory set is contradictory. 
And each step for proving a theory set $T=\left\{t_1, ..., t_n\right\}$ is contradictory can be represented as $\left(t_i, t_j, t_k\right)$, which means that the intermediate conclusion $t_k$ is generated based on $t_i$ and $t_j$ already existed in the theory set $T$ (intermediate conclusions generated by previous reasoning steps have been merged into the theory set $T$). Then, for a theory set $T$ with contradiction and one of its reasoning steps $\left\{t_i, t_j, t_k\right\}$, we can generate four training samples for training Pre-Selector, Post-Selector, and Knowledge Composer, respectively:

\begin{shrinkeq}{-3ex}
    \begin{align*}
    & Pre\text{-}S \ Input = \left\{T\right\}; Pre\text{-}S\ Output = \left\{t_i, t_j\right\}       \\
    & Post\text{-}S\ Input = \left\{T, t_i\right\}; Post\text{-}S\ Output = \left\{t_j\right\}      \\
    & Post\text{-}S\ Input = \left\{T, t_j\right\}; Post\text{-}S\ Output = \left\{t_i\right\}      \\
    & KC           \ Input = \left\{t_i, t_j\right\}; KC           \ Output = \left\{t_k\right\}    \\
    \end{align*}
\end{shrinkeq}
The generative knowledge composer can learn the resolution rule implicitly after training by language modeling loss. The pre-selector and post-selector are classification models, so that their output is converted to class labels instead of text. We use binary cross entropy loss to train these two modules.

To train the verifier, we utilize the output of XLNET in post-selector, specifically the vector representation corresponding to the [SEP] token via a linear layer, as the vector representations of the theories for simplicity. So the verifier and post-selector are trained jointly, with their loss function combined using a hyperparameter $\alpha$. For the example in Figure \ref{fig:model}, $T=\left\{t_1, ..., t_n\right\}$ is the current theory set,  $V=\left\{v_1, ..., v_n\right\}$ is the corresponding vector representations, $t_n$ is the theory selected by the pre-selector. Assume that $P=\left\{p_1, ..., p_k\right\}$ represents the indices of theories that can form a valid theory pair with $t_n$, which are considered as positive examples, $R=\left\{r_1, ..., r_m\right\}$ represents the indices of theories that cannot, which are considered as negative examples. The specific definition of the validity contrastive loss (VCE) is shown as follows, where the maximum similarity between positive examples is constrained to be 0.8 to prevent model collapse:
\begin{shrinkeq}{-1ex}
$$L_{vce}=-\frac{1}{k}\sum_{j=1}^{k}log{\frac{exp(max(sim(v_n,v_{p_j}),0.8))}{\sum_{i=1}^{m}{exp(sim(v_n,v_{r_i}))}}}$$
\end{shrinkeq}

\vspace{-0.3ex}
\section{Experiment Setup}
\vspace{-0.2ex}
\textbf{Tasks and Datasets}  
Following \citet{richardson2022pushing}, we trained and evaluated on the easy Ruletaker-depth-3ext dataset \cite{tafjord2021proofwriter}, then tested on the test set of Ruletaker-depth-3ext and Ruletaker-depth-5 dataset (later we will refer to them as Ruletaker-3ext and Ruletaker-D5) as well as the dev set of Hard Ruletaker (Hard RuleTaker only have the dev set). \textbf{Hard Ruletaker} is a \textbf{harder} dataset by eliminating potential bias \cite{2023ICML} compared to Ruletaker-3ext and Ruletaker-D5. 
However, the Hard Ruletaker dataset only includes True and False labels without Unknown labels, which may not accurately reflect the ability of the model. So we use the same method to sample hard instances whose label is Unknown and added it to Hard Ruletaker dataset to balance three kinds of labels. We named the new dataset \textbf{Hard Ruletaker*}. 
To further evaluate the performance of our model after training on hard instances, we divide Hard Ruletaker* to train, dev, and test set based on a ratio of 8.5,0.5,1, the divided dataset was called \textbf{Hard Ruletaker**}. 
However, these datasets do not contain existence quantifiers that are implicitly expressed in natural language, so we also use a method similar to \citet{richardson2022pushing} to construct a dataset with existence quantifiers called \textbf{Ruletaker-E}. 

To train GFaiR, we first get each data's FOL representation and then employ a resolution refutation based FOL prover to automatically derive the intermediate reasoning process. Finally, we transform it from FOL representations into natural language representations by using natural language templates. More details can be seen in Appendix \ref{sec:dataset}.

\noindent
\textbf{Baselines} For the first task, we compare GFaiR with two kinds of methods:

(1) \textbf{Pretrained Language Model Based Methods:} We use Roberta-large \cite{liu2019roberta}, T5-large and ChatGPT (gpt-3.5-turbo) as baselines. For Roberta-large and T5-large, we finetune them on the Ruletaker-3ext and Hard Ruletaker** datasets. For ChatGPT, we use the method of instruct and chain-of-thought prompt to evaluate its performance. Due to cost reasons, we respectively tested 3000 pieces of data in three datasets.
\begin{table}[!htbp] 
\centering
\small
\setlength{\tabcolsep}{1.5mm}{
\begin{tabular}{c c c c c c c c c c} 
\toprule 
\multicolumn{1}{c}{\multirow{2}{*}{Model}}& \multicolumn{2}{c}{Ruletaker-3ext}& \multicolumn{2}{c}{Hard RT}& \multicolumn{2}{c}{Hard RT*}\\
\cmidrule(r){2-3}\cmidrule(r){4-5}\cmidrule(r){6-7}
\multicolumn{1}{c}{}&EA&FA &EA&FA &EA&FA  \\ 
\hline 
\specialrule{0em}{1.5pt}{1.5pt}
\multicolumn{1}{c}{T5}        &97.7& —   &57.3& —   &57.5& —      \\
\multicolumn{1}{c}{Roberta}   &98.9& —   &59.6& —   &59.7& —      \\
\multicolumn{1}{c}{ChatGPT}   &56.5&42.8 &57.0&2.7  &38.9&6.9     \\
\multicolumn{1}{c}{IBR}       &98.9&98.1 &59.6&12.1 &59.7&29.6    \\
\multicolumn{1}{c}{FaiRR}     &99.0&98.4 &14.1&12.2 &41.1&39.8    \\
\multicolumn{1}{c}{NLProofs}  &\textbf{99.3}&\textbf{99.2} &14.3&13.8 &41.8&41.4 \\
\multicolumn{1}{c}{\textbf{GFaiR}}  &98.1&98.0 &\textbf{68.5}&\textbf{67.5} &\textbf{73.9}&\textbf{71.7}    \\ 
\bottomrule 
\end{tabular}
}
\caption{Comparison of GFaiR with baselines when trained on Ruletaker-3ext and tested on Ruletaker-3ext and two hard datasets. EA, FA, Hard RT, and Hard RT* refer to entailment accuracy, full accuracy, Hard Ruletaker, and Hard Ruletaker* respectively.}
\label{tab:generalize}
\end{table}

(2) \textbf{Stepwise Inference Methods}: we mainly compare GFaiR with the model combined with forward chaining FaiRR \cite{sanyal2022fairr} and the model combined with backward chaining IBR \cite{qu2022interpretable}. And we also compare GFaiR with NLProofs \cite{yang2022generating} which conducts proof search on partial proof graphs. More details can be seen in Appendix \ref{sec:baselines}.

\noindent
\textbf{Evaluation protocol} Following \citet{qu2022interpretable}, We consider two main aspects for evaluating the model's performance in our study: (1) \textbf{Entailment accuracy (EA)} measures how accurately the model is able to predict the label of the hypothesis. 
(2) \textbf{Full accuracy (FA)} measures how accurately the model can simultaneously predict the label and the valid proof (i.e. the reasoning process) of the hypothesis. For a reasoning process $P = (p_1, p_2… p_n)$, it is valid if and only if every reasoning step $p_i$ is correct. A reasoning step $p_i$ includes selected rules or facts $s_i$, along with reasoning conclusion $c_i$. To check if $p_i$ is right, we use the FOL format expression of $s_i$ and $c_i$, denoting as ${fs}_i$ and ${fc}_i$. And we consider $p_i$ is right if ${fc}_i$ can be directly derived by ${fs}_i$ using a valid reasoning rule under FOL. Following \citet{tafjord2021proofwriter}, when the model predicts Unknown, no proof will be generated and we think the proof is right when the gold label is Unknown. Note that our method for evaluating the reasoning process is more flexible compared to the evaluation method proposed in previous work \cite{saha2020prover}, which relies on precisely matching between the gold proof and the predicted proof. Instead, our evaluation method is able to take different reasoning paths into account. However, our method still will not evaluate incorrect reasoning processes as correct ones ensured by symbolized logical reasoning. 

\section{Experiment Results} 

\subsection{Main results}

\begin{table}[!htbp] 
\centering
\small
\begin{tabular}{c c c c c c c c c c} 
\toprule 
\multicolumn{1}{c}{\multirow{2}{*}{depth}}& \multicolumn{2}{c}{FaiRR}& \multicolumn{2}{c}{NLProofs}& \multicolumn{2}{c}{GFaiR}\\
\cmidrule(r){2-3}\cmidrule(r){4-5}\cmidrule(r){6-7} 
\multicolumn{1}{c}{}&EA&FA &EA&FA &EA&FA\\  
\hline 
\specialrule{0em}{1.5pt}{1.5pt}
\multicolumn{1}{c}{N/A}  &99.4&99.4 &99.4&99.4 &96.2&96.2    \\
\multicolumn{1}{c}{0}    &100 &100  &100 &100  &99.9&99.9     \\
\multicolumn{1}{c}{1}    &99.5&99.2 &99.9&99.9 &99.5&99.5    \\
\multicolumn{1}{c}{2}    &98.4&96.0 &99.0&99.0 &98.2&97.9    \\
\multicolumn{1}{c}{3}    &93.1&84.8 &94.1&93.4 &95.8&95.1    \\
\multicolumn{1}{c}{4}    &88.8&77.3 &79.5&77.2 &94.2&92.5    \\
\multicolumn{1}{c}{5}    &78.7&67.8 &69.6&57.3 &94.2&91.9    \\ 
\bottomrule 
\end{tabular}
\caption{Depth-wise results on Ruletaker-D5, N/A represents the depth is unknown because the value of the hypothesis is ‘unknown’.}
\label{tab:D5}
\end{table}

To investigate different methods' in-domain performance on easy problems and zero-shot generalization ability on hard problems, we trained and evaluated on easy Ruletaker-3ext dataset, and then tested on Ruletaker-3ext and two hard datasets (Hard Ruletaker and Hard Ruletaker*). Results are shown in Table \ref{tab:generalize}, from which we can observe that:

(1) Compared to pretrained language model based methods (T5-large, Roberta-large, and ChatGPT), we can find that stepwise inference methods are more faithful than ChatGPT from the difference between the value of EA and FA. 

(2) Compared to stepwise inference methods IBR, FaiRR, and NLProofs, GFaiR shows comparable performance on the biased RuleTaker-3ext dataset, and significantly outperforms on two debiased hard datasets, which demonstrates stronger zero-shot generalization performance according to EA and FA. This suggests that by introducing resolution refutation to improve completeness, the stepwise inference methods can generalize to complex logical reasoning scenarios. In contrast, previous stepwise inference methods IBR, FaiRR, and NLProofs are incomplete and often classify hypotheses that can be inferred as True or False as Unknown, so they exhibit unsatisfied when dealing with complex logical reasoning scenarios. Hence GFaiR's ability to generalize to complex logical reasoning scenarios is better.

(3) From the difference between the value of EA and FA, we can observe that our model is faithful. Though this difference of FaiRR as well as GFaiR on two hard datasets is much smaller, their entailment accuracy is relatively low so there is no point in only considering their faithfulness. However, GFaiR both achieves higher entailment accuracy and maintains faithfulness by combining resolution refutation and introducing a validity contrastive loss-based verifier. 

\begin{table}[!htbp] 
\centering
    \small
    \begin{tabular}{c c c c c} 
    \toprule 
    \multicolumn{1}{c}{\multirow{2}{*}{Model}}& \multicolumn{2}{c}{Hard RuleTaker**}&  \multicolumn{2}{c}{RuleTaker-E}\\
    \cmidrule(r){2-3}\cmidrule(r){4-5}
    \multicolumn{1}{c}{}&EA&FA &EA&FA\\  
    \hline 
    \specialrule{0em}{1.5pt}{1.5pt}
    \multicolumn{1}{c}{T5}              &87.1& —   &75.7& —         \\
    \multicolumn{1}{c}{Roberta}         &89.3& —   &76.8& —         \\
    \multicolumn{1}{c}{IBR}             &89.3&39.2 &76.8&35.3       \\
    \multicolumn{1}{c}{FaiRR}           &40.4&34.0 &38.4&36.6       \\
    \multicolumn{1}{c}{NLProofs}        &40.7&39.4 &38.6&38.2       \\
    \multicolumn{1}{c}{\textbf{GFaiR}}  &\textbf{92.2}&\textbf{92.2} &\textbf{83.2}&\textbf{82.7}    \\
    \bottomrule 
    \end{tabular}
    \caption{Results on Hard Ruletaker** and Ruletaker-E dataset.}
    \label{tab:hardexist}
\end{table} 

(4) ChatGPT does not outperform other models significantly and even performs worse than them. On the one hand, this reflects the difficulty of this task. On the other hand, this is because ChatGPT is a general-purpose model. However, the performance of some relatively small task-specific models far exceeds ChatGPT, demonstrating the immense potential of transformers in mastering logical operation rules and the necessity of equipping the data-driven chatGPT with the logical rules for enhancing the performance on complex rule reasoning tasks such as math or coding.  

Note that the EA of ChatGPT on Hard Ruletaker is slightly higher than on Ruletaker-3ext, this is because the labels in Hard Ruletaker are only True or False and we exclude data that ChatGPT considers Unknown (less than 10\%). Though this may overestimate the performance, it does not affect our conclusion. Additionally, the EA of IBR on hard datasets is much higher than FaiRR and equal to Roberta. This is because IBR first predicts the final answer and then gives a reasoning process, and only the reasoning process is derived by stepwise backward inference.

\subsection{Generalization to Higher Depths}
In this section, we experiment with a setting where models are trained on reasoning depths less than or equal to 3 and tested on Ruletaker-D5 which contains problems that require reasoning up to depth 5. The reasoning depth are defined based on the minimal reasoning depth using the forward-chaining reasoning 
method \cite{tafjord2021proofwriter}. But we use resolution refutation which is different from forward-chaining in principle and thus leads to different minimal reasoning depth for the same instance. However, in a statistical sense, data with higher reasoning depth for forward-chaining is generally higher for resolution refutation. So it can also be a reference to compare the generalization ability of different methods using the depth defined by previous work. 

From Table \ref{tab:D5}, we can find that the performance drop of GFaiR is smaller with the increasing reasoning depth. For example, considering the performance drops between d = 3 to d = 5, GFaiR has 1.6\% drop in entailment accuracy. In contrast, FaiRR and NLProofs drop 14.4\% and 24.5\% in entailment accuracy, respectively. 
\begin{table}[!htbp] 
    \centering
    \small
    \setlength{\tabcolsep}{2.8mm}{
    \begin{tabular}{c c c c c} 
        \toprule 
        \multicolumn{1}{c}{Model}  &5var&8var&10var&12var  \\  
        \hline 
        \specialrule{0em}{1.5pt}{1.5pt}
        \multicolumn{1}{c}{T5}   &95.5&87.8&82.3&80.9   \\
        \multicolumn{1}{c}{\textbf{GFaiR}}&\textbf{95.5}&\textbf{91.3}&\textbf{90.1}&\textbf{89.4}   \\   
        \bottomrule 
    \end{tabular}
    }
    \caption{Model performance on GRL dataset.}
    \label{tab:GRL}
\end{table}
\begin{table}[!htbp] 
    \centering
    \small
    \setlength{\tabcolsep}{2mm}{
    \begin{tabular}{c c c c c} 
        \toprule 
        \multicolumn{1}{c}{Model}  &16,21v&25,32v&35,48v&60,70v  \\  
        \hline 
        \specialrule{0em}{1.5pt}{1.5pt}
        \multicolumn{1}{c}{T5}   &88.2&87.4&82.9&77.4   \\
        \multicolumn{1}{c}{\textbf{GFaiR}}&\textbf{93.6}&\textbf{92.4}&\textbf{91.7}&\textbf{91.3}   \\
        \bottomrule 
    \end{tabular}
    }
    \caption{Model performance on RCL dataset.}
    \label{tab:RCL}
\end{table} 
This indicates that our model’s ability to generalize to higher reasoning depth is better.

\subsection{In-domain Performance on Complex Reasoning Scenarios}

To investigate the in-domain performance of different methods in complex reasoning scenarios, we evaluate different methods on Hard Ruletaker** dataset. Experimental results are shown in Table \ref{tab:hardexist}, from which we can find that compared to IBR, FaiRR, and NLProofs, GFaiR achieves better performance. Combining the experimental results in Table \ref{tab:generalize}, we can conclude that GFaiR is more effective in handling complex logical reasoning scenarios by introducing resolution refutation. 

\subsection{Performance on Ruletaker-E}
We also wish to see the performance on scenerios with implicitly expressed existence quantifiers. To do this, we evaluate different method's performances on the Ruletaker-E dataset. Experimental results are shown in Table \ref{tab:hardexist}, from which we can find that compared to FaiRR, IBR, and NLProofs, GFaiR achieves better performance, which indicates that it is also effective in handling implicitly expressed existence quantifiers by combining resolution refutation. Additionally, the difference between EA and FA also indicates that our model is faithful in scenarios with implicitly expressed existence quantifiers.

\subsection{Performance on Natural Language Satisfiability Task} 
\noindent
We further evaluate GFaiR on natural language satisfiability (NLSAT) task, whose aim is to determine whether there is a contradiction in the given NL Theory. In this task, we do not need the process of refutation, so the converter only needs to convert the NL Theory into natural language representations similar to the Skolem normal form, and then directly use our reasoning model to infer whether there is a contradiction in the given NL Theory. 

\begin{table}[!htbp] 
\centering
    \small
    \begin{tabular}{c c c c c c c} 
    \toprule 
    \multicolumn{1}{c}{\multirow{2}{*}{Model}}& \multicolumn{2}{c}{Ruletaker-3ext}&  \multicolumn{2}{c}{Hard Ruletaker*}\\
    \cmidrule(r){2-3}\cmidrule(r){4-5}
    \multicolumn{1}{c}{}&EA&FA &EA&FA \\  
    \hline 
    \specialrule{0em}{1.5pt}{1.5pt}
    \multicolumn{1}{l}{FaiRR}                                     &99.0&98.4 &41.1&39.8        \\
    \multicolumn{1}{l}{FaiRR+}                                    &98.4&98.3 &41.5&41.4        \\
    \multicolumn{1}{l}{GFaiR-}                                    &97.5&97.2 &72.4&68.6        \\
    \multicolumn{1}{l}{GFaiR}                                     &98.1&98.0 &73.9&71.7        \\
    \bottomrule 
    \end{tabular}
    \caption{Results of ablation study.}
    \label{tab:Ablation}
\end{table}
Specially, there are two datasets available in this task, Grounded Rule Language (GRL) dataset and Relative Clause Fragment (RCL) dataset. These datasets are more challenging compared to Hard RuleTaker \cite{richardson2022pushing}. This is because these datasets demand the model to reason only based on rules and the number 

of reasoning steps required to solve the problem significantly exceeds that of Hard Ruletaker. So we use these datasets to further investigate the performance of our approach in more complex reasoning scenarios. Because there are no facts available on these datasets and models designed with forward or backward chaining rely on facts during inference. Therefore, we cannot apply these models to such tasks. Instead, we compare GFaiR with the T5-large two-stage fine-tuning method \cite{richardson2022pushing}.

Experimental results are shown in Table \ref{tab:GRL} and \ref{tab:RCL}, from which we can observe that GFaiR outperforms the baseline methods on both datasets. Consequently, GFaiR is capable of handling more complex reasoning scenarios by combining the stepwise inference method and resolution refutation.

\subsection{Ablation Study}
To respectively explore the effects of resolution refutation and validity contrastive loss-based verifier in our model, we consider the following ablations: 1) FaiRR+: add the validity contrastive loss-based verifier to the FaiRR model. So comparing FaiRR+ with GFaiR can show the impact of resolution refutation; 2) GFaiR-: replace the validity contrastive loss-based verifier with the verifier proposed by \citet{yang2022generating} to check its impact.

Results on Ruletaker-3ext and Hard Ruletaker* datasets are given in table \ref{tab:Ablation}. From these results, we can know that even if adding a verifier to FaiRR, the performance on Hard Ruletaker* dataset is lower than GFaiR, which signifies the effectiveness of combining resolution refutation. Furthermore, we can know that GFaiR's performance is better than GFaiR-, which shows the effectiveness of the validity contrastive loss-based verifier. 

\section{Related Work}
\textbf{Natural Language Reasoning with First-Order Logic} First-order logic has a wide range of coverage. For example, it includes the majority of reasoning situations in commonsense reasoning \cite{davis2017logical}. Additionally, it can represent most problems in mathematics and domains such as Euclidean geometry, making it widely used in automated theorem provers \cite{nawaz2019survey}. As a result, FOL reasoning ability is a fundamental reasoning ability \cite{davis2017logical} widely used in existing reasoning benchmarks. For example, LogiQA \cite{liu2021logiqa} and ReClor \cite{Yu2020ReClor} are two benchmarks widely used in logical reasoning. 
However, \citet{tian2021diagnosing} points out that FOL reasoning ability is not disentangled from other reasoning abilities such as commonsense reasoning in 
these benchmarks. So even if a model performs poorly on these datasets, it can’t be concluded that the model lacks the reasoning ability. Starting from \citet{clark2021transformers}, there are a series of novel benchmarks which measure logical reasoning independently of other forms of reasoning. We focus on these benchmarks to check our model’s FOL reasoning ability. Since our method focuses on first-order logic reasoning based on natural language, it can easily be adapted to other forms of natural language-based reasoning problems.

\noindent
\textbf{Proof Generation} One of our task’s goals is to give a reasoning process, which is similar to the task of proof generation. Proof generation focuses on generating a reasoning chain from the given NL Theory to the conclusion, which aims at improving the model’s interpretability \cite{rudin2019stop,hase2020evaluating}. Recently, some works have been working on the problem of proof generation. Prover \cite{saha2020prover} trains a RoBERTa-based model that predicts nodes and edges of the proof graph. ProofWriter \cite{tafjord2021proofwriter} is a T5-based model that iteratively generates one-hop conclusions and proofs from the NL Theory. FaiRR \cite{sanyal2022fairr} further decomposes each reasoning step into selecting rules, selecting facts and reasoning based on selected rules and facts, which is similar to the reasoning process of forward reasoning. IBR \cite{qu2022interpretable} draws inspiration from backward reasoning and designs an iterative backward reasoning model. NU \cite{picco2021neural} also employs backward reasoning but it cannot generate a reasoning process. NLProofs \cite{yang2022generating} is also a stepwise reasoning method that using verifier-guided search. However, the validity contrastive loss-based verifier is more suitable for the reasoning scenerios of resolution. Another work MultiProver \cite{saha2021multiprover} aims at generating multiple proofs for a hypothesis. 

\section{Conclusion}
In this paper, we propose GFaiR, a faithful and generalizable model capable of handling complex logical reasoning scenarios by introducing a validity contrastive loss-based verifier and resolution refutation. 
Experimental results also shows that GFaiR achieves better performance especially on Hard RuleTaker and Hard RuleTaker* datasets.

\section{Acknowledgments}
We thank the anonymous reviewers for their constructive comments and gratefully acknowledge the National Natural Science Foundation of China (U22B2059, 62176079), and the Natural Science Foundation of Heilongjiang Province (Y02022F005).

\section{Bibliographical References}\label{sec:reference}

\bibliographystyle{lrec-coling2024-natbib}
\bibliography{lrec-coling2024-example}

\appendix
\section{Dataset details}
\label{sec:dataset}
Due to the need for intermediate reasoning processes when training our model, we employed the
FOL prover provided in the Stanford CS221 course page and Prover9 to automatically extract the reasoning process for these two types of tasks respectively. Below we describe our method to extract the reasoning process. Since the dataset is a synthetic dataset, regular expressions can be used to convert each data back to its corresponding FOL representation. Then we apply Prover9 or the FOL prover provided in the Stanford CS221 course page to each data and obtain its intermediate reasoning process of FOL representations. Finally, we transform the intermediate reasoning process from FOL representations into natural language representations by using natural language templates. Although this approach introduces some noise limited by the prover we
used (redundant and excessively long reasoning steps), it does not hinder our model from achieving excellent generalization performance across all tasks.

Additionally, \citet{richardson2022pushing} illustrated that they found around 1\% mismatched labels on the Ruletaker-3ext dataset. However, they only correct the train and dev set of the Ruletaker-3ext-sat dataset. As a result, we correct the test set of the Ruletaker-3ext dataset and the Ruletaker-D5 dataset for our experiments using the same method as \citet{richardson2022pushing}. 
\section{Baselines details}
\label{sec:baselines}
\subsection{ChatGPT Baseline}
In order to automatically evaluate the accuracy of the reasoning process generated by ChatGPT, we
need to know that the intermediate conclusions are derived from which theories are selected from
the theory set. Therefore, we use the form of instruct+chain-of-thought prompt to strictly restrict
its output form, specifically, we use 4-shot for Hard Ruletaker and 5-shot for Ruletaker-3ext-sat and Hard Ruletaker* because there is no label Unknown in Hard Ruletaker and we need one more example for the condition Unknown when testing on other datasets. However, there will still be a small portion of data (less than 10\%) that we cannot parse the output of ChatGPT, so we exclude this portion of data. We only tested 3000 pieces of data in three datasets respectively using gpt-3.5-turbo due to cost reasons.

\subsection{IBR Baseline}
Since IBR targets the problems in the Close World Assumption, we made simple modifications to
adapt to our target task. Specifically, the QA prediction module still first predicts the answer but we remove the strategy prediction module along with the strategy loss. This is because the search space of our target tasks is infinite so we can not generate proof when the answer is Unknown and the strategy is always Proof. As a result, if the QA prediction module predicts Unknown, we stop and return the results. On the contrary, we will apply other modules in IBR to get the reasoning process. During training, all the modules of IBR are trained together with three types of losses: parent prediction loss, child prediction loss, and QA prediction loss (strategic prediction loss has been removed). However, when the gold answer is Unknown, there is no parent prediction loss and child prediction loss, which will introduce some noise. As a result,
we implement the QA prediction module apart from two other modules.

\section{implementation details}
\label{sec:implementation}
To reduce the search space and improve the inference efficiency of our model, we combine two complete inference strategies specifically designed for resolution refutation when experimenting with the natural language reasoning with first-order logic task, including set of support strategy and linear resolution strategy. However, when experimenting with the natural language satisfiability task, we can not combine these inference strategies because the task is different. In addition, we use beam search with beam size 5 for RuleTaker-E and 2 for other datasets.

Previous work \cite{buss1998introduction} has shown that using linear resolution strategy and set of support strategy together will not affect the completeness of resolution refutation under first order logic. Set of support strategy requires that at least one of the two clauses involved in each resolution step is the negation of the inference target (hypothesis or the negation of the hypothesis) or a descendant of the negation of the inference target. Linear resolution strategy requires that one of the two clauses involved in each resolution step (except the first step) is the clause derived from the previous resolution step. Combining these two strategies, we can know that one of the two clauses involved in the first step is the negation of the inference target (from the set of support strategy), and one of the two clauses involved in the other steps is the clause derived from the previous resolution step (from the linear resolution strategy). From these we can know that one of the two clauses involved in each resolution step is determined. So we can remove the xlnet-based pre-selector while regarding that the pre-selector always choose the negation of the inference target in the first step, and choose the clause derived from the previous resolution step in the other resolution steps.  
 \end{document}